\documentclass[a4paper,11pt]{article}

\usepackage[a4paper,left=2.5cm,top=2.2cm,right=2.5cm,bottom=2.5cm]{geometry}
\usepackage[english]{babel}
\usepackage[T1]{fontenc}
\usepackage[utf8]{inputenc} 

\usepackage[protrusion=true,expansion=false]{microtype}
\usepackage[doublespacing]{setspace}

\usepackage{graphicx}
\graphicspath{{figures/}}           
\usepackage{float}
\usepackage{booktabs,multicol,multirow,tabularx}
\usepackage{subcaption}             
\usepackage{caption}
\usepackage{sidecap}

\usepackage{amsmath,amsthm,amssymb}
\usepackage{bm} 
\usepackage{mathtools}
\usepackage{dsfont}
\usepackage{bm}

\usepackage{diagbox}

\usepackage[dvipsnames,table]{xcolor}

\DeclarePairedDelimiter{\abs}{\lvert}{\rvert}

\DeclareMathOperator*{\argmin}{arg\,min}

\DeclareMathOperator*{\argminstop}{arg\,min\,stop}

\usepackage{enumitem}
\setlist[enumerate]{nosep}
\setlist[itemize]{nosep}

\usepackage{authblk}
\usepackage[hidelinks]{hyperref}
\usepackage[capitalize,nameinlink]{cleveref}

\usepackage{algorithm}
\usepackage{algpseudocode}

\usepackage{soul}
\sethlcolor{yellow}

\numberwithin{equation}{section}
\numberwithin{figure}{section}
\numberwithin{table}{section}

\newtheorem{theorem}{Theorem}[section]

\newcommand*{\R}{\mathds{R}}

\newcommand{\ew}{\mathbb{E}}

\newcommand{\emd}{\operatorname{EMD}}

\newcommand{\set}[1]{\left\{#1\right\}}

\newcommand{\source}{f}
\newcommand{\dbd}{g}

\newcommand{\zz}{\mathbf{z}}

\newcommand{\Vo}{\mathcal{V}}
\newcommand{\Uo}{\mathcal{U}}
\newcommand{\Ao}{\mathcal{A}}
\newcommand{\Po}{\mathcal{P}}
\newcommand{\Co}{\mathcal{C}}

\newcommand{\rr}{\mathbf{r}}
\newcommand{\yy}{\mathbf{r}_0}

\newcommand{\nn}{\xi}

\newcommand{\Ad}{{\bm{A}}}
\newcommand{\Bd}{{\bm{B}}}
\newcommand{\Cd}{{\bm{C}}}
\newcommand{\Pd}{{\bm{P}}}
\newcommand{\Vd}{{\bm{V}}}
\newcommand{\Ud}{{\bm{U}}}

\newcommand{\bd}{{\bm{g}}}
\newcommand{\bx}{{\bm{x}}}
\newcommand{\by}{{\bm{y}}}
\newcommand{\bX}{{\bm{X}}}
\newcommand{\bY}{{\bm{Y}}}
\newcommand{\bnoisea}{{\bm{\xi}}}
\newcommand{\bnoiseb}{{\bm{\eta}}}

\newcommand{\noisea}{\xi}

\newcommand{\invkernel}{\Phi}

\usepackage{siunitx}
\allowdisplaybreaks

\usepackage{pifont} 

\newenvironment{tritemize}
  {\begin{itemize}[
      label=\textcolor{green!30!gray}{\raisebox{0.2ex}{$\blacktriangleright$}},
      itemsep=0.0ex,
      topsep=0.0ex,
      parsep=0pt,
      partopsep=0pt,
      leftmargin=2em
    ]}%
  {\end{itemize}}
  
\title{Self-Supervised Angular Deblurring in Photoacoustic Reconstruction via Noisier2Inverse}

\author{Markus Haltmeier}
\affil{Department of Mathematics, University of Innsbruck\\
Technikerstrasse 13, 6020 Innsbruck, Austria\\
E-mail: \texttt{markus.haltmeier@uibk.ac.at}}
\author{Nadja Gruber}
\affil{Department of Computer Science, University of Innsbruck\\
Technikerstrasse 21a, 6020 Innsbruck, Austria\\
E-mail: \texttt{Nadja.Gruber@uibk.ac.at}}
\author{Gyeongha Hwang}
\affil{Department of Mathematics, Yeungnam University\\
280 Daehak-Ro, Gyeongsan, Gyeongbuk 38541, South Korea\\
E-mail: \texttt{ghhwang@yu.ac.kr}}
\date{\today}

\begin{document}

\maketitle

\begin{abstract}
Photoacoustic tomography (PAT) is an emerging imaging modality that combines the complementary strengths of optical contrast and ultrasonic resolution. A central task is image reconstruction, where measured acoustic signals are used to recover the initial pressure distribution. For ideal point-like or line-like detectors, several efficient and fast reconstruction algorithms exist, including Fourier methods, filtered backprojection, and time reversal. However, when applied to data acquired with finite-size detectors, these methods yield systematically blurred images. Although sharper images can be obtained by compensating for finite-detector effects, supervised learning approaches typically require ground-truth images that may not be available in practice. We propose a self-supervised reconstruction method based on Noisier2Inverse that addresses finite-size detector effects without requiring ground-truth data. Our approach operates directly on noisy measurements and learns to recover high-quality PAT images in a ground-truth-free manner. Its key components are: (i) PAT-specific modeling that recasts the problem as angular deblurring; (ii) a Noisier2Inverse formulation in the polar domain that leverages the known angular point-spread function; and (iii) a novel, statistically grounded early-stopping rule. In experiments, the proposed method consistently outperforms alternative approaches that do not use supervised data and achieves performance close to supervised benchmarks, while remaining practical for real acquisitions with finite-size detectors.

\medskip


\medskip

\end{abstract}

\section{Introduction}

In photoacoustic tomography (PAT), a target object subjected to a non-ionizing laser pulse absorbs the energy, undergoes thermoelastic expansion, and subsequently emits acoustic waves. The acoustic waves are measured by detectors placed around the object, from which the initial pressure distribution is reconstructed \cite{wang2015photoacoustic,li2009photoacoustic,rosenthal2013acoustic,kuchment2008mathematics}. Taking the finite size of the detectors into account, the PAT image reconstruction problem can be formulated as
\begin{equation}\label{eq:ip-pat}
\dbd
= \Uo_a(\source) + \nn,
\end{equation}
where $\source$ is the source term, $\Uo_a$ denotes the noise-free forward operator with $a$ representing detector aperture effects, and $\nn$ is additive noise. The data $\dbd$ corresponds to noisy measurements. The photoacoustic problem is of relevance in three spatial dimensions \cite{xu2003analytic,poudel2019survey}, typically using small or near point-like finite-area detectors, as well as in two spatial dimensions, where linear or line-like detectors are commonly used \cite{burgholzer2005thermoacoustic,paltauf2009characterization}. 

In the idealized setting of assuming point-like (3D) or line-like (2D) detectors, the initial pressure distribution can be recovered through efficient methods, including Fourier and series-based methods \cite{kunyansky2007series,xu2002exact,haltmeier2007thermoacoustic,kostli2001optoacoustic,kostli2001temporal,jaeger2007fourier}, time reversal techniques \cite{nguyen2016dissipative,burgholzer2007exact,treeby2010k,hristova2008reconstruction,stefanov2009thermoacoustic,stefanov2015multiwave}, or filtered back-projection-type inversion formulas \cite{finch2004determining,xu2005universal,kunyansky2007explicit,finch2007inversion,fawcett1985inversion,haltmeier2014universal,natterer2012photo,palamodov2012uniform}. However, in practical applications, several factors influence the reconstruction that are not included in the idealized model. These include wave propagation effects such as acoustic attenuation \cite{la2006image,burgholzer2007compensation,treeby2009fast,kowar2011causality,ammari2012photoacoustic}, as well as measurement-related limitations including finite sampling \cite{xu2002exact,haltmeier2016sampling} and detector size \cite{xu2003analytic,haltmeier2010spatial}. While all these effects are relevant, in this work we focus on the effect of finite detector size. We consider the 2D setting with linear detectors in circular scanning geometry. Extensions to point-like finite-area detectors will be discussed at several places in the manuscript.

PAT image reconstruction with the non-ideal model can be solved iteratively using matrix-model-based reconstruction algorithms \cite{wang2010imaging,dean2022practical}. However, these methods require significant computational resources due to the large matrices involved, and thus there is also interest in analytically simplified approaches. Using rotational invariance, in \cite{haltmeier2010spatial,roitner2014deblurring}, it is shown that for circular geometry this results in particular angular blurring. This observation serves as the basis for a two-step approach, where first the inverse of the idealized point-like model is applied and subsequently the resulting ill-posed angular deblurring problem is addressed. Related methods have been developed in several works \cite{roitner2014deblurring,rajendran2020deep,rejesh2013deconvolution,qi2021photoacoustic} to improve resolution limited by detector size. However, none of these methods employ learning-based self-supervised, ground-truth-free approaches.

Learning-based reconstruction methods achieve state-of-the-art performance in a wide range of inverse imaging problems, from medical imaging to computational photography~\cite{xu2014deep,jin2017deep,arridge2019solving,haltmeier2023regularization}. Most of these approaches rely on supervised learning, requiring paired data of noisy, blurred measurements and clean, sharp signals. However, collecting such ground-truth data can be costly or even unattainable, especially in medical and scientific imaging. Very recently, self-supervised methods have been proposed for denoising \cite{batson2019noise2self,moran2020noisier2noise}, undersampled Fourier inverse problems \cite{yaman2020self,millard2023theoretical,blumenthal2024self}, and computed tomography \cite{hendriksen2020noise2inverse,unal2024proj2proj,gruber2024sparse2inverse,gruber2025noisier2inverse,schut2026equivariance2inverse}. In this work, we introduce a self-supervised learning method for angular deblurring in photoacoustic tomography that does not require prior knowledge of the source. We establish a self-supervised framework, based on Noisier2Inverse \cite{gruber2025noisier2inverse}, for solving \eqref{eq:ip-pat}. Note that the application of the inverse of the idealized point-like model renders the noise correlated, in which case Noisier2Inverse is the natural choice. At its core, after a change to polar coordinates, the proposed approach reduces to a deconvolution problem addressed via self-supervised learning. To the best of our knowledge, no prior work has applied self-supervised learning in this context.

{
\color{black}
We conclude the introduction by summarizing our main contributions:

\begin{tritemize}
\item \textit{PAT-specific modeling:} We exploit the angular shift-invariance induced by finite detector aperture to recast PAT reconstruction as an angular deblurring problem in polar coordinates. Concretely, after applying the PAT inverses for point- and line-detector models together with Cartesian–polar transforms, the data model becomes a circular convolution in the angular variable, which we solve in the polar domain.

\item \textit{Noisier2Inverse in the polar domain:} We introduce a self-supervised deconvolution framework based on Noisier2Inverse that operates directly on the polar grid and leverages the known angular point-spread function. It eliminates the need for clean ground truth while preserving acquisition physics via circular angular convolutions. Notably, transforming \eqref{eq:ip-pat} into a circular convolution via a nonunitary transform induces correlated noise; therefore, methods that rely on (approximately) uncorrelated/i.i.d. noise—such as Sparse2Inverse—cannot be directly applied, whereas Noisier2Inverse does not suffer from this restriction.

\item \textit{Novel early-stopping rule:} As in $Y$-loss–based splitting methods, early stopping is essential because the data-space loss alone cannot control components in the (large) null space and those amplified by small singular values (ill-conditioning) of the forward operator. We propose a statistically grounded rule that matches the empirical distribution of predicted residuals to the given noise distribution, yielding an automated, validation-free stopping mechanism.
\end{tritemize}
}

\section{Background}

\subsection{Photoacoustic Tomography}

We consider PAT in two spatial dimensions, which is relevant when measuring data using integrating line detectors \cite{burgholzer2005thermoacoustic,paltauf2017piezoelectric,bauer2017all}. Acoustic wave propagation in this case is governed by the initial value problem for the 2D wave equation
\begin{alignat}{2}\label{eq:wave1}
\partial_{t}^2 u \left( \rr, t \right)  - \Delta u \left( \rr, t \right) &= 0   \quad && ~\text{for}~  (\rr, t )  \in \R^2  \times   (0, \infty ) \,,\\ \label{eq:wave2}
u    \left( \rr , 0 \right) &=  \source (\rr)  \quad   &&~\text{for} ~  \rr  \in \R^2  \,, \\ \label{eq:wave3}
\partial_t u\left(\rr, 0 \right)     &= 0           \quad              &&~\text{for} ~  \rr  \in \R^2 \,,
\end{alignat}
where $\source \colon \R^2 \to \R$ is the initial source, assumed to be supported in the unit disc $D = \set{\rr \in \R^2 \colon \abs{\rr} < 1}$. In an ideal setting, PAT consists of reconstructing the initial data $\source$ from Dirichlet boundary measurements
\begin{equation} \label{eq:wave4}
\Uo  \source   \triangleq  u|_{ \partial D \times (0,\infty)}  \,, 
\end{equation}
where $u|_{\partial D \times (0,\infty)}$ denotes the restriction of the unique solution of \eqref{eq:wave1} to the unit circle $\partial D$. Inversion of $\Uo$ is well studied. In particular, the operator $\Uo$ is well known to be invertible, with explicit stability estimates and range conditions available \cite{finch2006range,finch2007spherical,agranovsky2009range}. Moreover, explicit inversion formulas are well known and have been derived in \cite{kunyansky2007explicit,finch2007inversion,haltmeier2013inversion}.

In practice, data can only be measured up to a finite time, and truncating inversion formulas using signals for all times introduces artefacts. This issue has been addressed in \cite[Theorem 3.3]{dreier2022photoacoustic}, where an inversion formula for finite $T$ has been derived that will be used in our work.

\begin{theorem}[Finite-time inversion formula] \label{thm:invT}
Let $\source \colon \R^2 \to \R$ be a smooth function with compact support in $D$, and let $T>0$. Then, for every $\rr \in D$, we have
\begin{equation}\label{eq:inv}
f(\rr) 
\coloneqq  
 \frac{2}{\pi^2} \, \nabla_\rr \cdot \int_{S^1} \nu(\yy) \int_0^T  \invkernel_T(\|\rr-\yy\|,t)
 \frac{(\Uo \source) (\yy,t)}{ \sqrt{|\|\rr-\yy\|^2 - t^2|}} \, dt \, d \sigma(\yy) \,,
\end{equation} 
with  
\begin{equation}  \label{eq:inv-kernel}
\invkernel_T(r_1,r_2) 
:= 
\begin{cases}
\frac{1}{2} \log \left(\frac{\sqrt{T^2 - r_2^2}-\sqrt{r_1^2 - r_2^2}}{\sqrt{T^2 - r_2^2}+\sqrt{r_1^2 - r_2^2}}\right) & r_1>r_2 \,,\\
\arctan\left(\frac{\sqrt{T^2 - r_1^2}}{\sqrt{r_2^2 - r_1^2}}\right) & r_1<r_2 \,.
\end{cases}
\end{equation}
In particular, the expression on the right-hand side of \eqref{eq:inv} defines an explicit left inverse $\Vo$ of $\Uo$ using data in the finite time interval $[0,T]$.
\end{theorem}

\subsection{Finite-size detector blurring}

In this work we assume a more realistic scenario in which, instead of pointwise pressure data, pressure values integrated over a spatially extended detector aperture are measured. In circular geometry, and assuming a fixed detector surface profile independent of position, the measured data can be written as  \cite{xu2003analytic,poudel2019survey}
\begin{equation} \label{eq:wave5}
(\Uo_w \source)(\rr, t)
= \int_{S^1} w\!\bigl(\varphi(\rr,\yy)\bigr)\, (\Uo  \source)(\yy, t)\, d\yy \,,
\end{equation}
where $\varphi(\rr,\yy)$ denotes the angle between $\rr$ and $\yy$, and $w(\varphi)$ models the angular sensitivity profile of the detector. 

Using rotational invariance \cite{haltmeier2010spatial,roitner2014deblurring}, it has been shown that data \eqref{eq:wave5} for a source $\source$ corresponds, under the ideal inverse associated with \eqref{eq:wave4}, to data generated by an angularly blurred version of $\source$. This effect has been rigorously analyzed in two dimensions for a detector extending over a finite cylindrical aperture, which models, for instance, optical detection using a laser beam \cite{berer2012characterization} or an elongated optical fiber sensor \cite{paltauf2009characterization}. For later reference, we give a precise mathematical formulation of this result.

\begin{theorem}[Blurring due to finite detector size]\label{thm:blur}
Let $\mathbb{T} := [0,2\pi)$ with endpoints identified (period $2\pi$), and let $w \in L^1(\mathbb{T})$ be extended periodically to $\mathbb{R}$. Define, for $\source \colon \mathbb{R}^2 \to \mathbb{R}$ and $u \colon [0,\infty)\times \mathbb{T} \to \mathbb{R}$ that is $2\pi$-periodic in its angular variable,
\begin{align}
(\Po \source)(r,\phi) &= \source(r\cos\phi, r\sin\phi),\\
(\Ao_w u)(r,\phi) &= \frac{1}{2\pi}\int_{0}^{2\pi} u(r,\psi)\, w\big((\phi-\psi)\bmod 2\pi\big)\, d\psi,\\
(\Co u)(\rr) &=
\begin{cases}
u\bigl(\|\rr\|_2,\,\arg(\rr)\bigr), & \rr \neq 0,\\
u(0,0), & \rr = 0,
\end{cases}
\end{align}
where $\arg(\rr)\in\mathbb{T}$ is the principal argument. Note that $\Po$ and $\Co$ denote the change from Cartesian to polar coordinates and vice versa, respectively. Then, for any smooth function $\source \colon \mathbb{R}^2 \to \mathbb{R}$ with compact support in $D$, and for $\Uo$ and $\Uo_w$ defined by \eqref{eq:wave4} and \eqref{eq:wave5},  
\begin{equation}\label{eq:heinzi}
\forall\, \rr \in D:\qquad
\Uo_w \source(\rr) \;=\; \Uo\bigl(\Co \Ao_w \Po \source\bigr)(\rr).
\end{equation}
\end{theorem}

From Theorem \ref{thm:blur} we see that the data $\Uo_w \source$ obtained from an extended detector coincides with the ideal data $\Uo \source_a$, where the modified initial pressure is given by $\source_a = \Co \Ao_w \Po \source$. Hence, applying the inverse of the ideal forward operator to the data $\Uo_w \source$ from an extended detector leads to an angular deblurring problem, which we will treat using Noisier2Inverse.

\subsection{Noisier2Inverse}

In order to address angular deblurring, we make use of Noisier2Inverse \cite{gruber2025noisier2inverse}, a self-supervised reconstruction method for solving general inverse problems of the form
\begin{equation} \label{eq:ip}
\by = \Ad \bx + \bnoisea \,,
\end{equation}
where $\Ad \colon \bX \to \bY$ is a linear map between finite-dimensional spaces $\bX$ and $\bY$. It assumes a probabilistic setting in which the desired signal $\bx$ and the noise $\bnoisea$ are independent random vectors with distributions $p_{\bx}$ and $p_{\bnoisea}$, respectively, and the data $\by$ is generated according to \eqref{eq:ip}. 

The goal is to construct a reconstruction map $\Bd \colon \bX \to \bY$ based on well-defined selection criteria. In an ideal setting where the joint distribution of $\bx$ and $\by$ is known, one minimizes the supervised loss $\mathbb{E}\,\|\Bd \by - \bx\|^2$. This serves as the basis of supervised learning using samples $(\by^{(n)}, \bx^{(n)})$ of measurement data and corresponding ground-truth. However, such data are often difficult to obtain in practice. This has led to increasing interest self-supervised approaches, which define a reconstruction operator based on samples of $\by$ only. Such methods have been developed for denoising \cite{batson2019noise2self,moran2020noisier2noise}, magnetic resonance imaging \cite{batson2019noise2self,millard2023theoretical}, and more recently computed tomography \cite{hendriksen2020noise2inverse,gruber2024sparse2inverse,gruber2025noisier2inverse,unal2024proj2proj}. 

Noisier2Inverse \cite{gruber2025noisier2inverse} is a self-supervised method for solving \eqref{eq:ip}, applicable to general linear inverse problems with additive signal-independent noise. The basic principle of Noisier2Inverse is to add additional noise $\bnoiseb$ sampled from the same noise model $p_{\bnoisea}$ to the already noisy data $\by$, resulting in noisier data $\by + \bnoiseb$. Then a reconstruction network $\Bd \colon \bX \to \bY$ is trained such that $\Ad \Bd$ takes $\by + \bnoiseb$ as input and outputs an estimate of $\by - \bnoiseb$. More formally, $\Bd$ is defined as the minimizer of $\mathbb{E}\| \Ad \Bd(\by + \bnoiseb) - (\by - \bnoiseb) \|_2^2$. Mathematically, Noisier2Inverse is supported by the following theoretical result.
 
\begin{theorem}[Noisier2Inverse]\label{thm:nn2i}
Let $(\bX,\langle\cdot,\cdot\rangle_\bX)$ and $(\bY,\langle\cdot,\cdot\rangle_\bY)$ be finite-dimensional real Hilbert spaces with norms $\|\bx\|_\bX := \sqrt{\langle \bx,\bx\rangle_\bX}$ and $\|\by\|_\bY := \sqrt{\langle \by,\by\rangle_\bY}$. Let $\Ad \colon \bX \to \bY$ be a deterministic linear map, and let $\bx \in X$ and $\bnoisea,\bnoiseb \in Y$ be square-integrable random vectors such that $\bnoisea$ and $\bnoiseb$ are independent, identically distributed, and independent of $\bx$. Define $\by := \Ad\bx + \bnoisea$. Then
\begin{equation}\label{eq:nn2i}
\argmin_{\Bd} \; \ew \, \big\|\, \Ad\Bd(\by + \bnoiseb) - (\by - \bnoiseb) \,\big\|_\bY^2
\;=\;
\argmin_{\Bd} \; \ew \, \big\|\, \Ad\Bd(\by + \bnoiseb) - \Ad\bx \,\big\|_\bY^2,
\end{equation}
where the minimum is taken over all measurable functions $\Bd \colon \bY \to \bX$ such that $\Ad\Bd(\by + \bnoiseb)$ is square-integrable (so both risks in \eqref{eq:nn2i} are finite).
\end{theorem}

Let us summarize the key ingredients and contributions of Noisier2Inverse. First, inspired by Sparse2Inverse~\cite{gruber2024sparse2inverse}, a network $\Ad\Bd$ is trained in $\bY$-space while being factorized through the learned component $\Bd \colon \bX \to \bY$ for image reconstruction and the fixed forward operator $\Ad \colon \bY \to \bX$. Second, inspired by Noisier2Noise~\cite{moran2020noisier2noise}, the network $\Ad\Bd$ is trained on noisier data $\by + \bnoiseb$ to overcome the absence of ground-truth data. A direct application of Noisier2Noise would suggest minimizing $\ew \,\|\Ad\Bd(\by + \bnoiseb) - \by\|_\bY^2$. However, in this case the trained network only performs half of the desired denoising and requires an additional extrapolation step during inference. Thus, as a third component, Noisier2Inverse proposes to target $\by - \bnoiseb$, which avoids any extrapolation during inference.

\section{Self-Supervised angular deblurring framework}

We now turn to the solution of the inverse problem \eqref{eq:ip-pat} with noisy PAT data according to \eqref{eq:wave5}. Our strategy is to first reformulate \eqref{eq:ip-pat} as a deconvolution problem in the polar domain via \eqref{eq:heinzi}, and then apply Noisier2Inverse combined with a novel automated stopping criterion.

\subsection{Reconstruction strategy}

{\color{black}
According to Theorem~\ref{thm:blur} we have
$\Uo_w\,(\source) = \Uo\,\Co\,\Ao_w\,\Po\,(\source)$,
where $\Po$ maps a planar function to polar coordinates, $\Ao_w$ denotes angular convolution in the polar domain with kernel $w$, $\Co$ maps back to Cartesian coordinates, and $\Uo$ and $\Uo_w$ are the ideal and extended-detector forward operators, respectively. Since $\Vo$ is a left inverse of $\Uo$ and $\Po$ is a left inverse of $\Co$, applying $\Po\,\Vo$ yields
$\Po\,\Vo\,\Uo_w(\source) = \Ao_w\,\Po(\source)$.
Similarly, applying $\Po\,\Vo$ to \eqref{eq:ip-pat} gives
\begin{equation}\label{eq:dec}
\dbd_P \;=\; \Ao_w(\source_P) \;+\; \noisea_P,
\end{equation}
with data $\dbd_P \triangleq \Po\,\Vo(\dbd)$, noise $\noisea_P \triangleq \Po\,\Vo(\noisea)$, and unknown $\source_P \triangleq \Po(\source)$. Therefore, estimating $\source$ reduces to solving the deconvolution problem \eqref{eq:dec} in the polar domain.
}

Building upon \eqref{eq:dec} the initial data $\source$ can be reconstructed by the following steps.
\begin{enumerate}[label=(A\arabic*)]
\item \label{A1} Apply the left inverse $\Vo$ to get $\Vo(\dbd)$.
\item \label{A2} Change to polar coordinates to get $\dbd_P \triangleq \Po(\Vo(\dbd))$.
\item \label{A3} Compute an estimate $\hat\source_P$ by solving the deconvolution problem \eqref{eq:dec}.
\item \label{A4} Transform back to Cartesian coordinates: $\hat \source \triangleq \Co(\hat\source_P)$.
\end{enumerate}
In the numerical implementation we realize all steps in a discrete manner. To this end, we discretize all arising operators in steps \ref{A1}, \ref{A2}, \ref{A4}, and solve \ref{A3} algorithmically. The concrete discretizations $\Ud$, $\Vd$, $\Pd$, $\Cd$, and $\Ad$ of $\Uo$, $\Vo$, $\Po$, $\Co$, and $\Ao_w$ are not central at this stage and are described in the numerical Section \ref{sec:num_spec}. 
What is crucial here is that, after discretization, we can rewrite \eqref{eq:dec} as
\begin{equation}\label{eq:dec-dec}
\by_{\Pd} \;=\; \Ad\!\big(\bx_{\Pd}\big) \;+\; \bnoisea_{\Pd}\,,
\end{equation}
with a discrete convolution operator $\Ad \colon \bX \to \bY$. This yields a finite-dimensional linear inverse problem of the form \eqref{eq:ip}. In this paper, we solve \eqref{eq:dec-dec} using our self-supervised Noisier2Inverse approach.

\subsection{Application of  Nosier2Inverse}

In the following we assume a random model for \eqref{eq:dec-dec} in which the unknown $\bx_P$ and the noise $\by_P$ are independent random variables. Due to the application of $\Vd$, the noise becomes correlated, so techniques such as Sparse2Inverse cannot be applied directly. However, Noisier2Inverse is applicable to correlated noise and is therefore well suited to this setting.

Following the general strategy of Noise2Inverse, we add additional noise $ \bnoiseb_P$ to $\by_P$, where $ \bnoiseb_P$ follows the same noise model as $ \bnoisea_P$, and then minimize
\begin{equation}\label{eq:loss}
	\mathcal{L}(\Bd) =
	\ew \big[ \| (\Ad \Bd)[ \by_P + \bnoiseb_P ]
	- ( \by_P - \bnoiseb_P ) \|_\bY^2 \big].
\end{equation}
This results in a trained component $\Bd \colon \bY \to \bX$ that solves the deconvolution problem \eqref{eq:dec-dec}, alongside a fixed deterministic component $\Ad \colon \bX \to \bY$. From Theorem \ref{thm:nn2i}, it follows that \eqref{eq:loss} is equivalent to minimizing the supervised $\bY$-space loss $\ew \big[ \| \Ad \Bd( \by_P + \bnoiseb_P ) - \Ad \bx \|_\bY^2 \big]$.

In the practical realization, \eqref{eq:loss} is not strictly minimized. Instead, the reconstruction function $\Bd$ is chosen from a parameterized class $(\Bd(\cdot;\theta))_{\theta \in \Theta}$, and the empirical risk
\begin{equation}\label{eq:lossN}
\mathcal{L}_N^{\textnormal{NN2I}}(\theta) \triangleq
\frac{1}{N} \sum_{n=1}^N \| (\Ad \Bd)( \by_P^{(n)} + \bnoiseb^{(n)}_P ;\theta)
	- ( \by^{(n)}_P - \bnoiseb^{(n)}_P ) \|_\bY^2
\end{equation}
is minimized using samples $(\by_P^{(n)})_{n=1}^N$ and $(\bnoiseb_P^{(n)})_{n=1}^N$ of noisy data and additional noises, respectively. Moreover, to avoid overfitting, we do not fully minimize \eqref{eq:lossN}; instead, we apply an iterative algorithm to \eqref{eq:lossN} combined with a novel stopping rule, described in Section~\ref{sec:stop} below.

The overall reconstruction framework is summarized in Algorithm \ref{alg:recon}.

\begin{algorithm}[htb!]
\caption{Discrete reconstruction via polar deconvolution (Noisier2Inverse)}
\label{alg:recon}
\begin{algorithmic}[1]
\Require discrete data $\bd$; discrete operators $\Vd,\Pd,\Cd,\Ad$
\Require network architecture $\bigl(\Bd(\cdot;\theta)\bigr)_{\theta \in \Theta}$
\Require self-supervised loss $\mathcal{L}_N^{\textnormal{NN2I}}$ using data $\{(\by_P^{(n)}, \bnoiseb_P^{(n)})\}_{n=1}^N$
\Ensure reconstruction $\hat{\bx}$
\State $\theta^\star \gets \argminstop_{\theta}\, \mathcal{L}_N^{\textnormal{NN2I}}(\theta)$ \Comment{Self-supervised training (\eqref{eq:lossN}, early stopping)}
\State $\by \gets \Vd\,\bd$ \label{a1} \Comment{Apply discrete left inverse}
\State $\by_P \gets \Pd\,\by$ \label{a2} \Comment{Map to polar domain}
\State $\hat{\bx}_P \gets \Bd(\by_P;\theta^\star)$ \label{a3} \Comment{Deconvolution in polar domain}
\State $\hat{\bx} \gets \Cd\,\hat{\bx}_P$ \label{a4} \Comment{Back to Cartesian domain}
\State \Return $\hat{\bx}$
\end{algorithmic}
\end{algorithm}

\subsection{Stopping criterion}
\label{sec:stop}

In the numerical minimization of self-supervised learning, we apply iterative techniques to minimize \eqref{eq:lossN}. Specifically, we utilize the Adam optimizer \cite{kingma2014adam}. However, exact minimization suffers from overfitting. This arises from at least two sources: the finite amount of training data, which leads to the empirical risk \eqref{eq:lossN}, and, more fundamentally, the fact that the loss is formulated in $\bY$-space, thus inheriting the ill-posedness of the deconvolution problem. In practice, we observed that iterative minimization of \eqref{eq:lossN} produces sharp, clean images in the early training stages, whereas prolonged training leads to artifacts. This behavior reflects both the ill-posedness of the deconvolution and the use of a data-space loss, and it resembles classical issues in iterative methods for inverse problems \cite{kaltenbacher2008iterative}. To address this, we propose an early-stopping strategy.

The proposed stopping criterion compares, at each iteration, the distribution of the predicted noise with the specified noise distribution $p_{\bnoisea}$. We quantify the distributional difference by the Earth Mover’s Distance (EMD; the 1-Wasserstein distance). At iteration $k$, let $\{\by^{(n)}\}_{n=1}^N$ and $\{\bnoisea^{(n)}\}_{n=1}^N$ denote samples of data and noise from a validation dataset, and consider the predicted noise $\by^{(n)} - \Ad \Bd_k \by^{(n)}$, where $\Bd_k$ is the reconstruction operator at iteration $k$. We then quantify the distributional discrepancy between the empirical measures $\frac{1}{N}\sum_{n=1}^N \delta\big(\,(\cdot) - \bnoisea^{(n)}\big)$ and $\frac{1}{N}\sum_{n=1}^N \delta\big(\,(\cdot) - (\by^{(n)} - \Ad \Bd_k \by^{(n)}) \big)$ via the optimal transport problem
\begin{equation}\label{eq:emd}
\begin{aligned}
\emd_k \;=\; \min_{F \in \mathbb{R}^{N \times N}} \; & \sum_{n,m=1}^N
F_{n,m}\, \big\| \bnoisea^{(n)} - \big( \by^{(m)} - \Ad \Bd_k \by^{(m)} \big) \big\|_2
\\
&\text{s.t.} \;
\begin{cases}
F_{n,m} \ge 0, \\[2pt]
\sum_{m=1}^N F_{n,m} = \tfrac{1}{N}, \\[2pt]
\sum_{n=1}^N F_{n,m} = \tfrac{1}{N}.
\end{cases}
\end{aligned}
\end{equation}
Here, $F_{n,m}$ is the transportation plan (flow) from $\bnoisea^{(n)}$ to $\by^{(m)} - \Ad \Bd_k \by^{(m)}$. During training, $\emd_k$ is evaluated at each iteration, and early stopping is triggered when $\emd_k$ attains its minimum on the validation set. 

\begin{figure}[htb!]
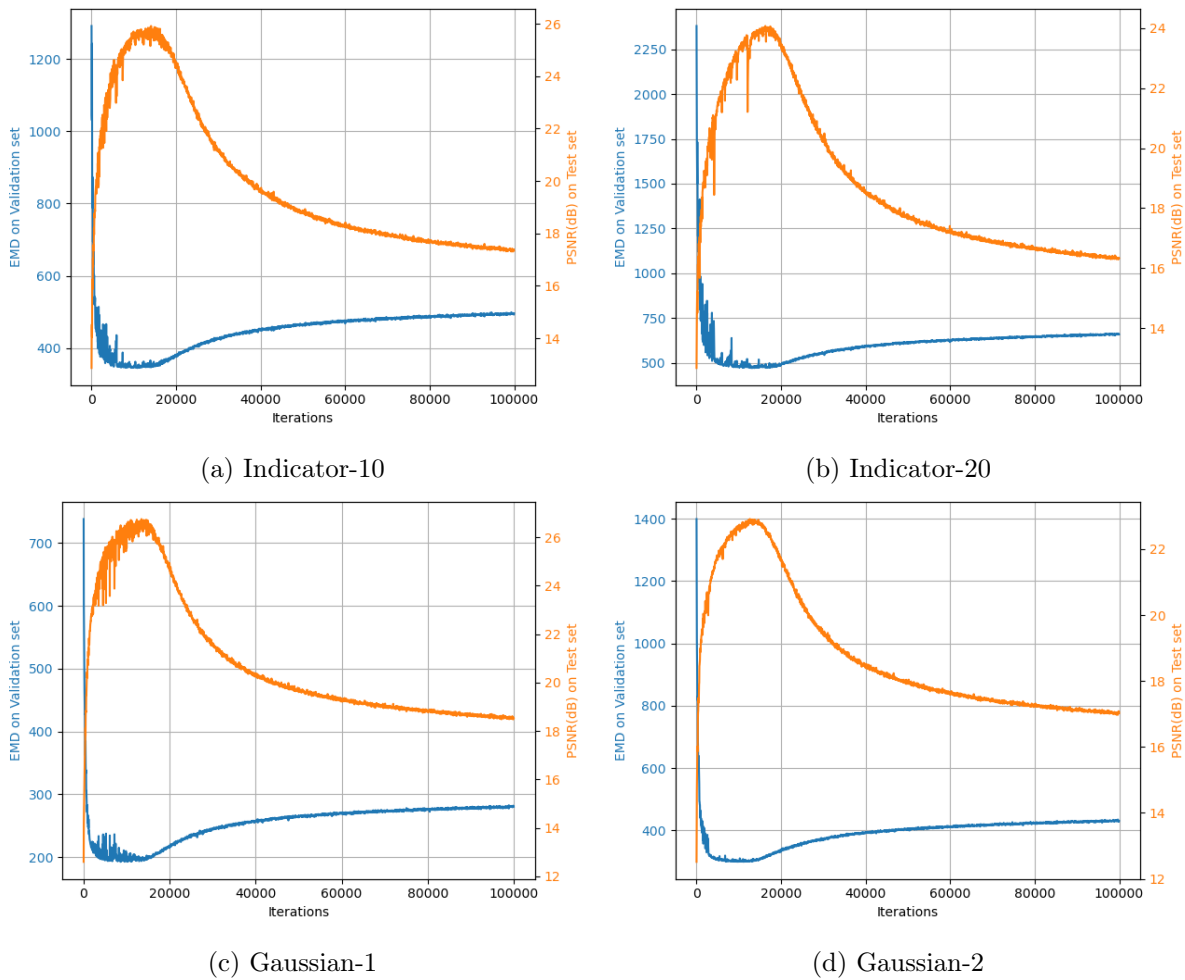

\centering
\captionsetup[subfigure]{justification=centering}
\newcommand{\panel}[2]{\includegraphics[width=\linewidth]{Earlystop/#1/#2/emd_Psnr_plot.png}}

\begin{subfigure}[t]{0.49\textwidth}
\centering
\panel{Indicator-10}{0.1}
\subcaption{Indicator-10}
\end{subfigure}
\begin{subfigure}[t]{0.49\textwidth}
\centering
\panel{Indicator-20}{0.1}
\subcaption{Indicator-20}
\end{subfigure}
\begin{subfigure}[t]{0.49\textwidth}
\centering
\panel{Gaussian-1}{0.1}
\subcaption{Gaussian-1}
\end{subfigure}
\begin{subfigure}[t]{0.49\textwidth}
\centering
\panel{Gaussian-2}{0.1}
\subcaption{Gaussian-2}
\end{subfigure}

\caption{Comparison between $\emd_k$ on the validation set (black) and PSNR at iteration $k$ on the test set (orange) for four different blurring kernels. All results consistently show an inverse relationship between these quantities.}
\label{fig:emd_Psnr}
\end{figure}

Numerical results reveal an inverse relationship between the validation-set EMD and the test-set PSNR, as illustrated in Fig.~\ref{fig:emd_Psnr} (implementation details are provided in the section below). Consequently, we employ early stopping that selects the iterate minimizing the validation-set EMD, thereby preventing overfitting at later iterations.   

\section{Numerical simulations}

In this section we present numerical experiments of the self-supervised discrete reconstruction framework outlined in Algorithm \ref{alg:recon}. All experiments were conducted in Python. To this end, we first specify the discretization, $\Ud$, $\Vd$, $\Pd$, $\Cd$, and $\Ad$, give implementation details for Noisier2Inverse, and then present results and discussion.

\subsection{Inverse problem specifications}
\label{sec:num_spec}

\paragraph{Forward and inverse wave models:}
The initial pressure $\source$ is discretized as $\bx$ on a $M \times M$ Cartesian grid over $[-1,1]^2$ and supported in the unit disc. For the results presented we use $M=256$. The ideal forward map is obtained by first solving \eqref{eq:wave1}--\eqref{eq:wave3} numerically via the FFT,
\begin{equation}\label{eq:sol-discrete}
p(\cdot,t_n)
= \operatorname{Re}\!\left[
\operatorname{FFT}^{-1}\!\Big(\operatorname{FFT}(\bx)\cdot \cos\big(t_n \sqrt{\kappa_1^2+\kappa_2^2}\big)\Big)
\right],
\end{equation}
where $(\kappa_1,\kappa_2)$ are the discrete frequency coordinates and the product is taken pointwise in the frequency domain. Samples according to \eqref{eq:wave4} on the detector circle are collected via bilinear interpolation. Detectors are placed on the unit circle at uniformly distributed angles $\phi_k \in [0,2\pi)$ with coordinates $(\cos\phi_k,\sin\phi_k)$. We use full temporal and angular sampling according to PAT sampling theory \cite{haltmeier2016sampling}, resulting in $N_{\textnormal{det}}  = \big\lfloor \pi N_x \big\rfloor = 804$ angular samples and $M$ temporal samples with $t_n = 2n/M$. Overall, each simulated sinogram has size $N_{\textnormal{det}}\times M$. This defines the discrete forward operator $\Ud \colon \mathbb{R}^{M\times M} \to \mathbb{R}^{N_{\textnormal{det}}\times M}$. The finite-time FBP inversion formula in implemented following \cite{dreier2022photoacoustic}, resulting in a map $\Vd \colon \mathbb{R}^{N_{\textnormal{det}}\times M} \to \mathbb{R}^{M\times M}$.

\paragraph{Blurred data and coordinate change:}
Observed noisy data are generated by discretizing \eqref{eq:heinzi}, resulting in $\Ud \Cd \Ad \Pd \bx + \bnoisea$ and subsequently adding additive Gaussian white noise with standard deviation $\alpha\,\|\Ud \Cd \Ad \Pd \bx\|_\infty$ for noise level $\alpha\in\{0.02,\,0.05,\,0.10\}$. Here we set $N_r = M/2$ and $N_\phi = \lfloor \pi M \rfloor$, and use the discrete switches to and from polar coordinates $\Pd \colon \mathbb{R}^{M\times M} \to \mathbb{R}^{N_\phi\times N_r}$ and $\Cd \colon \mathbb{R}^{N_\phi\times N_r} \to \mathbb{R}^{M\times M}$ with the Python package \texttt{polarTransform}. The operator $\Ad$ is the angular discrete convolution (circular along the angular coordinate). Discrete angular blurring is the linear map $\Ad:\mathbb{R}^{N_\phi\times N_r}\to \mathbb{R}^{N_\phi\times N_r}$ acting as a circular convolution along $\phi$; it is implemented in PyTorch via \texttt{torch.nn.functional.conv2d} with circular padding along the angular dimension.

\paragraph{Kernel specifications:}
We use periodic angular kernels (wrapped on $[0,2\pi)$) with a finite discrete support of length $63$ samples. The four kernels are:
\begin{tritemize}
\item \texttt{Indicator-10}
\item \texttt{Indicator-20}
\item \texttt{Gaussian-1}
\item \texttt{Gaussian-2}.
\end{tritemize}
Here \texttt{Indicator-10} and \texttt{Indicator-20} are indicator (box) kernels corresponding to detector apertures of $10^\circ$ and $20^\circ$. For an aperture $a$ (in degrees), the discrete half-width in samples is $n=\lfloor (a/720)\,N_\phi \rfloor$; the kernel is uniform on $\Delta\ell\in[-n,n]$, zero elsewhere, then periodized and normalized. The \texttt{Gaussian-1} and \texttt{Gaussian-2} kernels are (periodized) discrete Gaussians with $\sigma\in\{1,2\}$ angular-sample units: $h[\Delta\ell]\propto \exp(-(\Delta\ell/\sigma)^2/2)$, truncated to $63$ samples and normalized to sum to one.

\subsection{Network training and evaluation}

\paragraph{Data and preprocessing:}
For training and evaluation we use the FIVES dataset \cite{jin2022fives}, which contains 800 high-resolution color fundus photographs with corresponding retinal vessel segmentation masks. We focus on the vessel segmentation component; all images and masks are resized to $256\times256$ pixels. The split is $600$/$100$/$100$ for training/validation/testing. Given an image $\bx$, we form its polar representation via $\Pd$, apply angular blurring $\Ad$, and the forward operator $\Ud$ to obtain the training targets in the measurement domain. Noisier2Inverse is then actually performed in the polar image domain after applying $ \Pd \circ \Vd$ to the noisy measurement data.

\paragraph{Noisier2Inverse details:}
The network operates on polar grids of size $2M \times (M/2)$.
We use a U-Net–style encoder–decoder $\Bd(\theta; \cdot)$ with skip connections that maps polar inputs to polar outputs,
$ \Bd(\theta; \cdot) \colon \mathbb{R}^{2M \times (M/2)}\to\mathbb{R}^{2M \times (M/2)}$.
The architecture has four resolution levels, $3\times3$ convolutions with ReLU activations, and a final $1\times1$ convolution to the output channel. Circular padding is used along the angular dimension to respect periodicity in $\phi$. We minimize \eqref{eq:lossN} with Adam \cite{kingma2014adam} using a batch size of $15$, $10^{5}$ iterations, and learning rate $10^{-4}$. Model selection is based on the earth-mover distance based stopping criteria described in Section \ref{sec:stop} applied on the validation split. At test time we evaluate the trained network on $\by_P$, which empirically yields better reconstructions for our setting as applied to $\by_P + \bnoiseb_P$.

\paragraph{Comparison methods:}
For comparison, we replace Noisier2Inverse in Algorithm~\ref{alg:recon} with several learning-based deconvolution methods. All approaches use the same U-Net and a learning rate of $10^{-4}$. Specifically, we consider supervised learning, self-supervised learning with a discrete total variation penalty (SSLTV) \cite{zhang2021general}, and deep image prior (DIP) \cite{ulyanov2018deep}:
\begin{align}\label{eq:super}
\mathcal{L}_N^{\textnormal{sup}}(\theta)
&= \frac{1}{N}\sum_{i=1}^N 
\big\|\Bd\big(\by_P; \theta\big) - \bx_P\big\|_2^2 \,,\\
\label{eq:ssltv}
\mathcal{L}_N^{\textnormal{TV}}(\theta)
&= \frac{1}{N}\sum_{i=1}^N 
\big\|\Ad\,\Bd\big(\by_P; \theta\big) - \by_P\big\|_2^2
+ \lambda\,\big\|\Bd\big(\by_P; \theta\big)\big\|_{\mathrm{TV}} \,, \\
\label{eq:dip}
\mathcal{L}_{\by}^{\textnormal{DIP}}(\theta)
&= \big\|\Ad\,\Bd(\zz_P; \theta) - \by_P\big\|_2^2 \,,
\end{align}
where $\|\cdot\|_{\mathrm{TV}}$ denotes discrete total variation. Supervised learning uses paired data $(\bx^{(i)},\by^{(i)})$ as the gold standard. SSLTV requires no ground-truth images; $\lambda\in\{10^{-3},10^{-2},10^{-1},1\}$ is chosen by the best validation PSNR with early stopping on the validation set. DIP uses no training pairs; $\zz_P\in\mathbb{R}^{2M\times M}$ is a fixed random image (uniform) sized to the network domain; we run up to $4\times10^{4}$ iterations with early stopping by validation PSNR. While determining early stopping based on maximum PSNR on the test set is not realistic in practice, we report it to compare our method with the best results achievable by the baselines.

\definecolor{kindblack}{RGB}{198,219,239}
\definecolor{kindgreen}{RGB}{204,235,197}
\newcommand{\bestcell}[1]{{\cellcolor{kindgreen}{#1}}}
\newcommand{\secondcell}[1]{{\cellcolor{kindblack}{#1}}}

\begin{table}[htb!]
\centering
\small
\setlength{\tabcolsep}{4pt}
\sisetup{parse-numbers=false}
\begin{tabular}{l *{12}{S[table-format=2.2]}}
\toprule
& \multicolumn{3}{c}{\textbf{Indicator-10}}
& \multicolumn{3}{c}{\textbf{Indicator-20}}
& \multicolumn{3}{c}{\textbf{Gaussian-1}}
& \multicolumn{3}{c}{\textbf{Gaussian-2}}\\
\cmidrule(lr){2-4}\cmidrule(lr){5-7}\cmidrule(lr){8-10}\cmidrule(lr){11-13}
\textbf{Method}
& {$0.02$} & {$0.05$} & {$0.10$}
& {$0.02$} & {$0.05$} & {$0.10$}
& {$0.02$} & {$0.05$} & {$0.10$}
& {$0.02$} & {$0.05$} & {$0.10$}\\
\midrule 
DIP         & \secondcell{31.03} & \secondcell{26.62} & 23.48 & \bestcell{29.70} & 24.81 & 22.02 & \secondcell{29.06} & \secondcell{26.70} & \secondcell{24.35} & 24.14 & 22.78 & \secondcell{21.56} \\
SSLTV       & 29.99 & 26.25 & \secondcell{23.65} & 28.87 & \secondcell{25.06} & \secondcell{22.22} & 29.05 & 26.19 & 23.89 & \secondcell{24.21} & \secondcell{22.89} & 21.39 \\
Ours        & \bestcell{31.93} & \bestcell{28.48} & \bestcell{25.75} & \secondcell{29.54} & \bestcell{26.52} & \bestcell{23.81} & \bestcell{31.16} & \bestcell{28.39} & \bestcell{26.42} & \bestcell{25.69} & \bestcell{24.11} & \bestcell{22.80} \\
\bottomrule
\end{tabular}
\caption{PSNR (dB) between the reconstruction produced by each method from the observed test data $\bx^{(n)}$ and the inverse-projected clean, sharp oracle sinogram $(\Vd \circ \Ud)(\bx^{(n)})$. Best results among the unsupervised methods are highlighted with a \bestcell{green box}, second best with a \secondcell{black box}.}
\label{tab:psnr}
\end{table}

\subsection{Numerical results}

\begin{figure}[htb!]
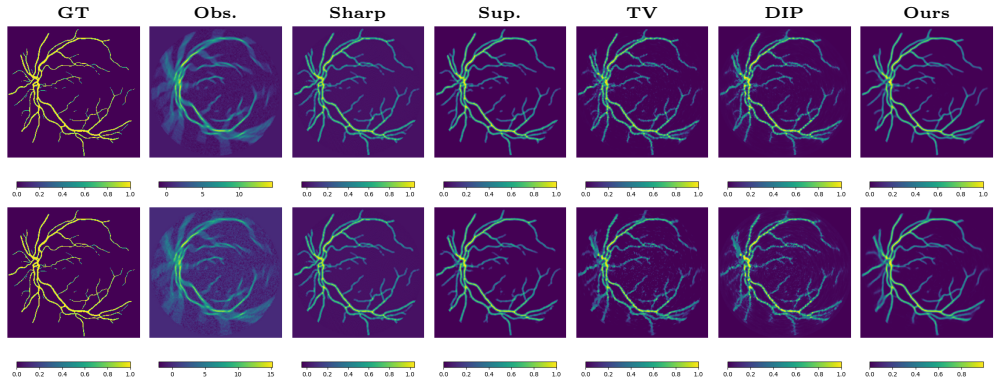
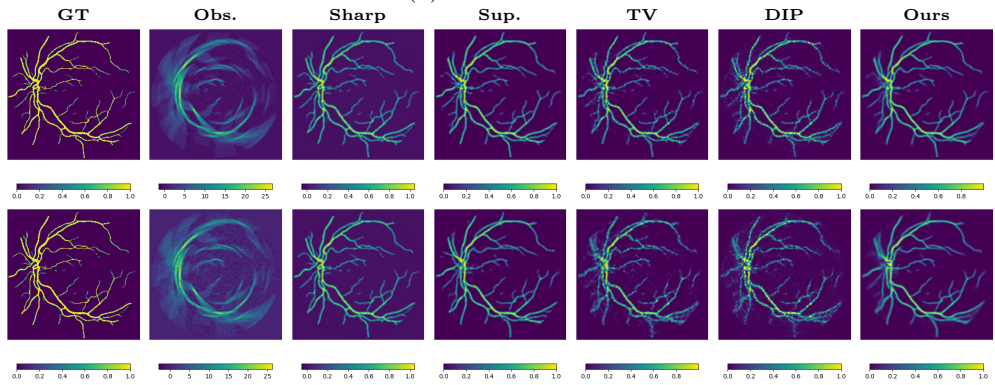
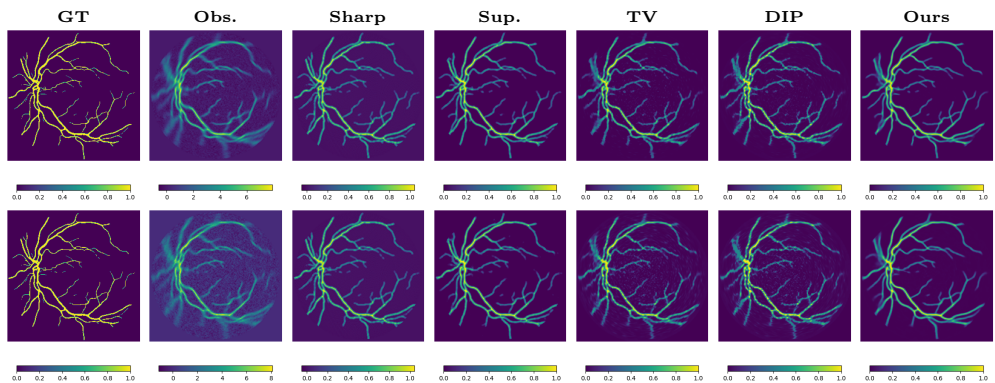
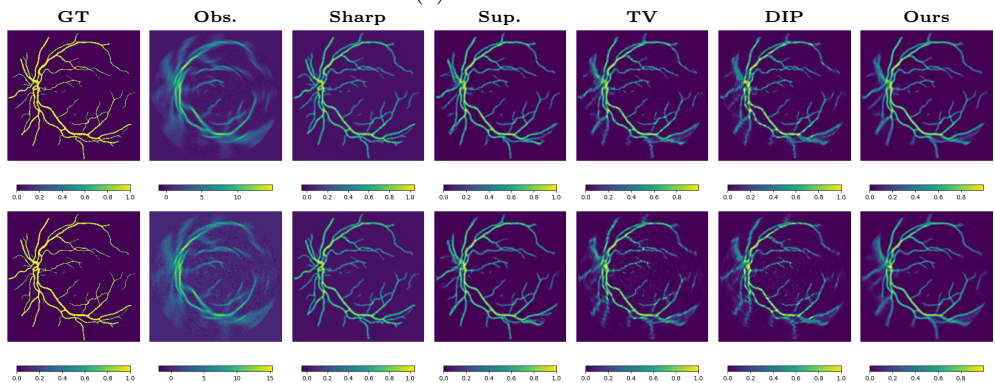

\centering
\captionsetup{skip=2pt}
\captionsetup[subfigure]{justification=centering}
\setlength{\tabcolsep}{0.5pt}
\renewcommand{\arraystretch}{0.88}

\newcommand{\imgw}{0.135\linewidth}
\newcommand{\img}[1]{\includegraphics[width=\imgw]{#1}}

\newcommand{\rowpics}[2]{%
  \img{Earlystop/#1/#2/00000000G.T..png}&
  \img{Earlystop/#1/#2/00000000Blurry.png}&
  \img{Earlystop/#1/#2/00000000Sharp.png}&
  \img{Earlystop/#1/#2/0best0000Test_S.png}&
  \img{Earlystop/#1/#2/0best0000Test_TV.png}&
  \img{Earlystop/#1/#2/Best0000Test_DIP.png}&
  \img{Earlystop/#1/#2/0best0000Test_U2.png}%
}

\begin{subfigure}[t]{0.85\textwidth}
\centering
\tiny
\begin{tabular}{@{}*{7}{c}@{}}
\textbf{GT} & \textbf{Obs.} & \textbf{Sharp} & \textbf{Sup.} & \textbf{TV} & \textbf{DIP} & \textbf{Ours} \\
\rowpics{Indicator-10}{0.05} \\[-1pt]
\rowpics{Indicator-10}{0.1}
\end{tabular}
\subcaption{Indicator-10}
\end{subfigure}\hfill
\begin{subfigure}[t]{0.85\textwidth}
\centering
\tiny
\begin{tabular}{@{}*{7}{c}@{}}
\textbf{GT} & \textbf{Obs.} & \textbf{Sharp} & \textbf{Sup.} & \textbf{TV} & \textbf{DIP} & \textbf{Ours} \\
\rowpics{Indicator-20}{0.05} \\[-1pt]
\rowpics{Indicator-20}{0.1}
\end{tabular}
\subcaption{Indicator-20}
\end{subfigure}

\vspace{2pt}

\begin{subfigure}[t]{0.85\textwidth}
\centering
\tiny
\begin{tabular}{@{}*{7}{c}@{}}
\textbf{GT} & \textbf{Obs.} & \textbf{Sharp} & \textbf{Sup.} & \textbf{TV} & \textbf{DIP} & \textbf{Ours} \\
\rowpics{Gaussian-1}{0.05} \\[-1pt]
\rowpics{Gaussian-1}{0.1}
\end{tabular}
\subcaption{Gaussian-1}
\end{subfigure}\hfill
\begin{subfigure}[t]{0.85\textwidth}
\centering
\tiny
\begin{tabular}{@{}*{7}{c}@{}}
\textbf{GT} & \textbf{Obs.} & \textbf{Sharp} & \textbf{Sup.} & \textbf{TV} & \textbf{DIP} & \textbf{Ours} \\
\rowpics{Gaussian-2}{0.05} \\[-1pt]
\rowpics{Gaussian-2}{0.1}
\end{tabular}
\subcaption{Gaussian-2}
\end{subfigure}

\caption{Example reconstruction across blur types (subfigures). Within each subfigure, the two rows correspond to noise levels $0.05$ (top) and $0.10$ (bottom).}
\label{fig:recon1}
\end{figure}

\begin{figure}[htb!]
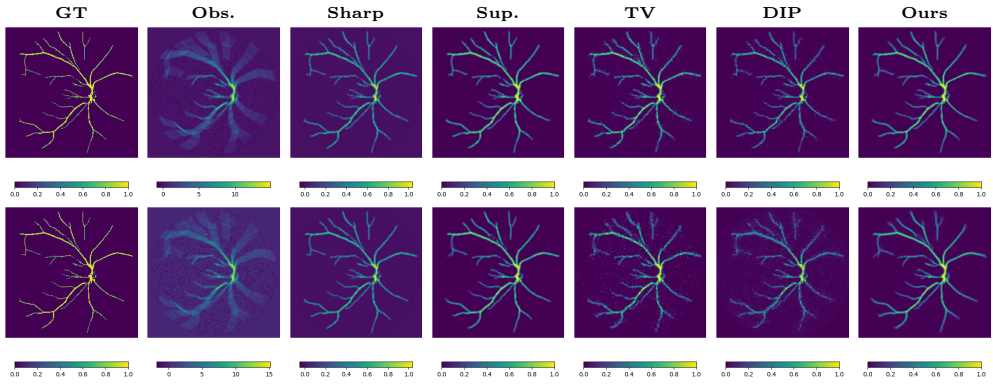
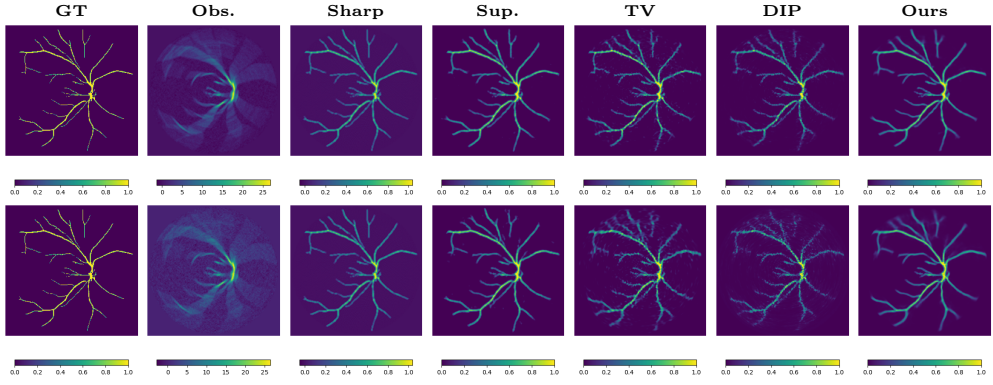
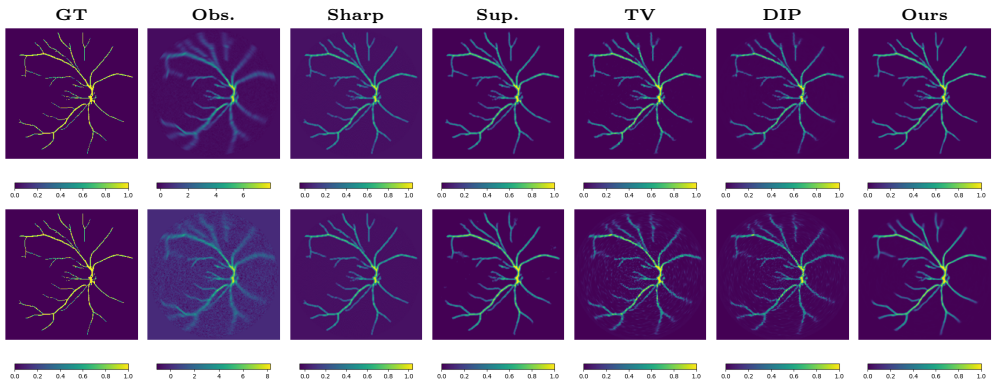
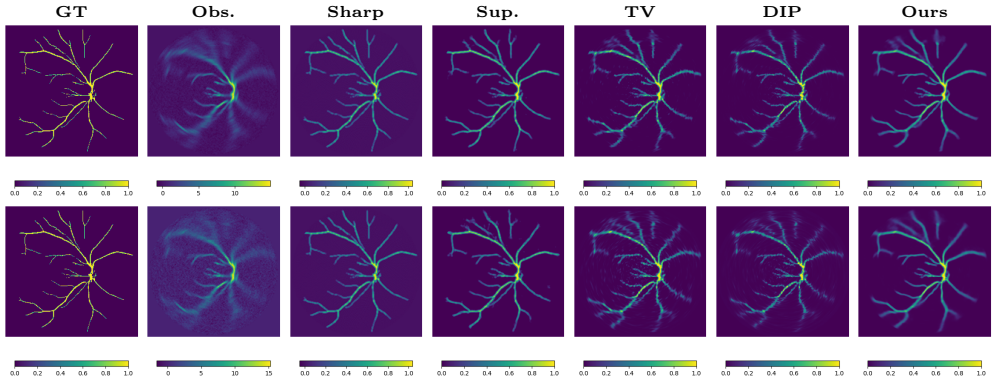

\centering
\captionsetup{skip=2pt}
\captionsetup[subfigure]{justification=centering}
\setlength{\tabcolsep}{0.5pt}
\renewcommand{\arraystretch}{0.88}

\newcommand{\imgw}{0.135\linewidth}
\newcommand{\img}[1]{\includegraphics[width=\imgw]{#1}}

\newcommand{\rowpicsB}[2]{%
  \img{Earlystop/#1/#2/00000001G.T..png}&
  \img{Earlystop/#1/#2/00000001Blurry.png}&
  \img{Earlystop/#1/#2/00000001Sharp.png}&
  \img{Earlystop/#1/#2/0Best0001Test_S.png}&
  \img{Earlystop/#1/#2/0Best0001Test_TV.png}&
  \img{Earlystop/#1/#2/Best0001Test_DIP.png}&
  \img{Earlystop/#1/#2/0Best0001Test_U2.png}%
}

\begin{subfigure}[t]{0.85\textwidth}
\centering
\tiny
\begin{tabular}{@{}*{7}{c}@{}}
\textbf{GT} & \textbf{Obs.} & \textbf{Sharp} & \textbf{Sup.} & \textbf{TV} & \textbf{DIP} & \textbf{Ours} \\
\rowpicsB{Indicator-10}{0.05} \\[-1pt]
\rowpicsB{Indicator-10}{0.1}
\end{tabular}
\subcaption{Indicator-10}
\end{subfigure}\hfill
\begin{subfigure}[t]{0.85\textwidth}
\centering
\tiny
\begin{tabular}{@{}*{7}{c}@{}}
\textbf{GT} & \textbf{Obs.} & \textbf{Sharp} & \textbf{Sup.} & \textbf{TV} & \textbf{DIP} & \textbf{Ours} \\
\rowpicsB{Indicator-20}{0.05} \\[-1pt]
\rowpicsB{Indicator-20}{0.1}
\end{tabular}
\subcaption{Indicator-20}
\end{subfigure}

\vspace{2pt}

\begin{subfigure}[t]{0.85\textwidth}
\centering
\tiny
\begin{tabular}{@{}*{7}{c}@{}}
\textbf{GT} & \textbf{Obs.} & \textbf{Sharp} & \textbf{Sup.} & \textbf{TV} & \textbf{DIP} & \textbf{Ours} \\
\rowpicsB{Gaussian-1}{0.02} \\[-1pt]
\rowpicsB{Gaussian-1}{0.1}
\end{tabular}
\subcaption{Gaussian-1}
\end{subfigure}\hfill
\begin{subfigure}[t]{0.85\textwidth}
\centering
\tiny
\begin{tabular}{@{}*{7}{c}@{}}
\textbf{GT} & \textbf{Obs.} & \textbf{Sharp} & \textbf{Sup.} & \textbf{TV} & \textbf{DIP} & \textbf{Ours} \\
\rowpicsB{Gaussian-2}{0.05} \\[-1pt]
\rowpicsB{Gaussian-2}{0.1}
\end{tabular}
\subcaption{Gaussian-2}
\end{subfigure}

\caption{Same setup as Fig.~\ref{fig:recon1}, with a different ground-truth sample.}
\label{fig:recon2}
\end{figure}

We evaluate Noisier2Inverse and the reference methods in terms of PSNR (dB). PSNR is computed between each method’s reconstruction on the test set and the oracle inverse $\Vd\!\circ\!\Ud(\bx^{(n)})$ of the ground truth $\bx^{(n)}$. Results for the four blurring kernels are reported in Table~\ref{tab:psnr}. The proposed method outperforms the other unsupervised baselines, even though the baselines use oracle early stopping while our method uses an automated stopping rule. Example results for all methods and all kernels at the tested noise levels are shown in Figs.~\ref{fig:recon1} and \ref{fig:recon2}. These images display the ground truth $\bx^{(n)}$, the angularly blurred, time-reversed data $\Vd(\by^{(n)})$, the sharp oracle inverse $\Vd\!\circ\!\Ud(\bx^{(n)})$, and the reconstructions from the supervised and unsupervised methods. Visual inspection confirms that Noisier2Inverse yields the best unsupervised reconstructions and approaches the supervised benchmark.

\section{Conclusion and outlook}

Photoacoustic signal formation is governed by the 3D acoustic wave equation. With infinitely long line detectors, image reconstruction can be reduced to inverting a 2D wave equation for line-integrated pressures. In practice, however, (piezoelectric) detectors have a finite angular aperture, so the idealized 2D model is blurred in the angular coordinate, degrading resolution. Focusing on the circular geometry, we formulate this as an angular deblurring problem in polar coordinates and propose a self-supervised learning approach (Noisier2Inverse) inspired by Noisier2Inverse. Our method operates directly in the polar domain, requires no clean ground truth, and in experiments outperforms other self-supervised baselines while approaching supervised performance.

Future work will include evaluation on real data; broader comparisons with learning and non-learning baselines; incorporating additional resolution-limiting factors (e.g., finite bandwidth, sparse sampling); extending to other acquisition geometries; studying deconvolution in measurement space; benchmarking against methods such as Sparse2Inverse; and developing a theoretical analysis of the proposed automated stopping rule.

\section*{Acknowledgment}

The work of G. Hwang has been supported by the 2025 Yeungnam University Research Grant.

\bibliographystyle{unsrt}
\bibliography{ref-spin}

\end{document}